\documentclass[10pt,twocolumn,letterpaper]{article}

\usepackage{cvpr}              

\usepackage[accsupp]{axessibility}
\usepackage{comment}
\usepackage{color}
        
\usepackage[table]{xcolor}
\usepackage{booktabs}
\usepackage{enumitem}
\usepackage{graphicx}
\usepackage{amsmath}
\usepackage{booktabs}
\usepackage{mathtools}
\usepackage{url}
\usepackage{comment}
\usepackage{color}
\usepackage{graphicx,setspace}
\usepackage{multirow, array, booktabs,makecell,xspace,bm}
\usepackage{pifont}
\usepackage{amsfonts}
\usepackage{dsfont}
\usepackage{amssymb}
\usepackage{tcolorbox}
\usepackage{enumitem}
\usepackage{graphicx}
\usepackage{amsmath}
\usepackage{booktabs}
\usepackage{mathtools}
\usepackage{url}
\usepackage{comment}
\usepackage{color}
\usepackage{graphicx,setspace}
\usepackage{multirow, array, booktabs,makecell,xspace,bm}
\usepackage{pifont}
\usepackage{amsfonts}
\usepackage{arydshln}
\usepackage{dsfont}
\usepackage{multirow}

\usepackage{stackengine}

\usepackage{scalerel}
\usepackage{esdiff} 

\usepackage{bbm}

\usepackage{algorithm}
\usepackage{algpseudocode}
\algnewcommand{\LineComment}[1]{\State \# #1} 

\usepackage{tablefootnote}

\definecolor{cvprblue}{rgb}{0.21,0.49,0.74}
\usepackage[pagebackref,breaklinks,colorlinks,allcolors=cvprblue]{hyperref}

\def\paperID{****} 
\def\confName{CVPR}
\def\confYear{2026}

\title{Learning from Oblivion: Predicting Knowledge-Overflowed Weights \\ via Retrodiction of Forgetting}
\author{Jinhyeok Jang$^{1,2}$\qquad
Jaehong Kim$^{1}$ \qquad
Jung Uk Kim$^{3}$\footnotemark[1]\thanks{Corresponding author}\\
$^{1}$ETRI \qquad
$^{2}$UST \qquad
$^{3}$Kyung Hee University \\
{\tt\small \{jjh6297, jhkim504\}@etri.re.kr,  ju.kim@khu.ac.kr}
}

\begin{document}
\maketitle
\begin{abstract}
Pre-trained weights have become a cornerstone of modern deep learning, enabling efficient knowledge transfer and improving downstream task performance, especially in data-scarce scenarios. However, a fundamental question remains: how can we obtain better pre-trained weights that encapsulate more knowledge beyond the given dataset? In this work, we introduce \textbf{KNowledge-Overflowed Weights (KNOW)} prediction, a novel strategy that leverages structured forgetting and its inversion to synthesize knowledge-enriched weights. Our key insight is that sequential fine-tuning on progressively downsized datasets induces a structured forgetting process, which can be modeled and reversed to recover knowledge as if trained on a larger dataset. We construct a dataset of weight transitions governed by this controlled forgetting and employ meta-learning to model weight prediction effectively. Specifically, our \textbf{KNowledge-Overflowed Weights Nowcaster (KNOWN)} acts as a hyper-model that learns the general evolution of weights and predicts enhanced weights with improved generalization. Extensive experiments across diverse datasets and architectures demonstrate that KNOW prediction consistently outperforms Na\"ive fine-tuning and simple weight prediction, leading to superior downstream performance. Our work provides a new perspective on reinterpreting forgetting dynamics to push the limits of knowledge transfer. The code and pre-trained model are available at \href{https://github.com/jjh6297/KNOW}{https://github.com/jjh6297/KNOW}.
\end{abstract}  

\section{Introduction}

The use of pre-trained weights has become a fundamental aspect of deep learning, driving significant advancements in various computer vision tasks. These weights serve as essential initializations that capture rich and reusable representations learned from large-scale datasets, facilitating faster convergence and improving task-specific performance 
\cite{huh2016makes,kornblith2019better,raffel2020exploring,kim2021robust,abnar2021exploring}. In particular, fine-tuning the pre-trained weights has shown substantial benefits in scenarios with limited labeled data, such as few-shot learning \cite{he2019rethinking}. In such cases, pre-trained weights act as a reservoir of knowledge, reducing dependence on large amounts of labeled data and extensive computational resources.

\begin{figure}[t]
\centering
\begin{minipage}[b]{1.0\linewidth}
\centering
\centerline{\includegraphics[width=1.0\linewidth]{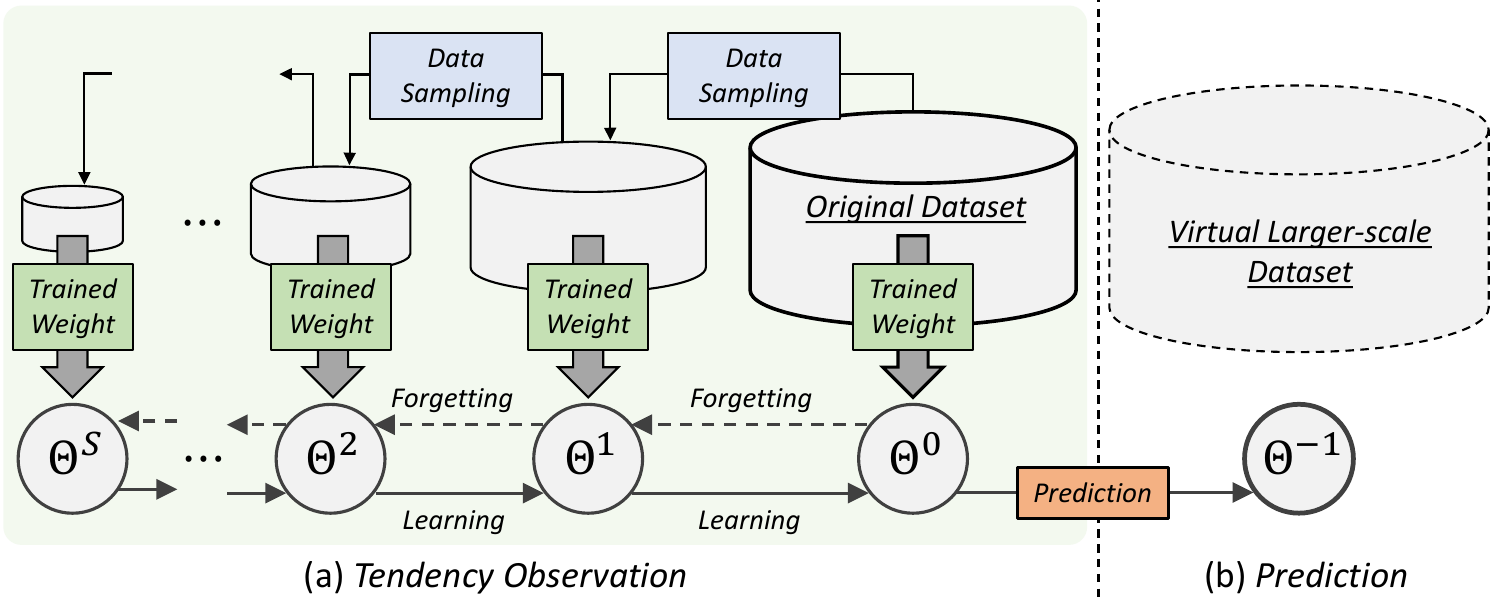}}
\end{minipage}
\caption{A conceptual diagram of the proposed task, \textbf{KNOW prediction}. We hypothesize the existence of a bidirectional relationship between progressive finetuning and forgetting and leverage this relationship to predict weights that encapsulate more knowledge than what is present in the given training dataset.}
\label{fig1}
\vspace{-0.2cm}
\end{figure}

At this point, we raise the two fundamental questions:

\noindent\rule[1pt]{\linewidth}{0.4pt}
\begin{enumerate}[label=\emph{(\roman*)}]
    \item \emph{What defines a better pre-trained weight?}
    \item \emph{If practical constraints exist, how can we obtain the better weights?}
\end{enumerate}
\noindent\rule[7pt]{\linewidth}{0.4pt}

For the question \textit{(i)}, a potential answer can be found in prior analyses about \textbf{scaling law} \cite{sun2017revisiting,kaplan2020scaling}, which suggest that increasing the size of the pre-training dataset generally leads to better pre-trained weights, thereby improving downstream task performance. However, as highlighted in question \textit{(ii)}, preparing such large datasets is often difficult in practice—for example, large-scale data collection and curation incur substantial cost and effort. Thus, it becomes essential to explore methods that can address this limitation and obtain the best possible pre-trained weights despite these constraints.

In this paper, instead of increasing the dataset size, we aim to emulate weights as if they had been trained on a scaled-up dataset. We introduce \underline{KN}owledge \underline{O}verflowed \underline{W}eights (\textbf{KNOW}) prediction, a novel strategy that predicts knowledge-enriched weights beyond those obtained from the original training set. As shown in Fig. \ref{fig1}, our approach intentionally induces sequential forgetting through a series of fine-tuning steps on progressively downsized subsets of the training dataset. By utilizing weight transitions from sequential fine-tuning, KNOW prediction enables us to reverse the forgetting trajectory and expect weights trained on a larger dataset than the original training set. Unlike conventional approaches, our method \textbf{extrapolates weights that retain and enhance prior knowledge}, serving as a knowledge-enriched initialization that accelerates convergence and improves downstream performance. This work provides new insights into how inverting the forgetting process can transform pre-trained weights into more knowledgeable and high-performing initializations.

For our KNOW prediction approach, we adopt meta-learning scheme. Meta-learning, often referred to as ``learning to learn," is a paradigm leveraging pre-acquired meta knowledge for efficient training \cite{andrychowicz2016learning, jang2023learning, knyazev2024accelerating}. Among various meta-learning frameworks, model-based meta-learning has gained significant attention, particularly in the weight space, where approaches such as weight prediction have been explored. These methods typically employ a hyper-model to predict the weights of a target model, adjusting them to a more suitable state to enhance training efficiency.

To develop our meta-learned model, we constructed a dataset of weight transitions structured by controlled sequential forgetting. We then employed meta-learning to enhance weight prediction. Specifically, our meta-learned model, \underline{KN}owledge \underline{O}verflowed \underline{W}eights \underline{N}owcaster (\textbf{KNOWN}), functions as a hyper-model that meta-learns the general tendencies of weight evolution, enabling more accurate KNOW prediction.

Our contributions can be summarized as follows: 
\begin{itemize}[leftmargin=2em]
\item  We introduce KNOW prediction, a novel strategy that leverages structured sequential forgetting and its inversion to synthesize weights with overflowed knowledge.

\item  We construct a dataset of weight transitions obtained through progressive forgetting, facilitating effective modeling of the forgetting trajectory.
    
\item  We propose KNOWN, a meta-hypermodel for predicting the weights trained on larger datasets.

\item  Extensive experiments across diverse datasets and architectures demonstrate consistent improvements, validating the effectiveness of virtual weights.

\end{itemize}

\section{Related Works}
\label{sec:formatting}

\subsection{Learning for Weight Prediction}
Research in areas such as Loss Landscape \cite{li2018visualizing}, Weight Averaging \cite{matena2022merging, ilharco2022patching, wortsman2022model, jang2025model}, Model Soup \cite{wortsman2022model, jang2025model}, and Task Vectors \cite{ilharco2022editing} suggests that the weight-loss surface around the converged weights of DNNs is relatively smooth. Based on this smoothness, it becomes possible to predict weights for specific purpose, such as efficient training.

\noindent \textbf{Weight Update Prediction for Training Acceleration} involves leveraging meta-learning to understand and predict how model weights evolve during training. A prominent approach, \textit{Learning to Optimize (L2O)} \cite{andrychowicz2016learning, lv2017learning, wichrowska2017learned, l2oamalgam}, replaces traditional optimizers with DNN-based optimizers designed to learn from prior training experiences. By predicting the better updates, L2O aims to accelerate convergence and improve training efficiency.
Other methods focus on forecasting future weight states to skip unnecessary training steps. Introspection \cite{sinha2017introspectionaccelerating} pioneered weight prediction by learning trajectories of model weights and identifying those that could be anticipated multiple epochs ahead, thus reducing training time. Building on this, the Weight Nowcasting Network (WNN) \cite{jang2023learning} introduced periodic nowcasting, using a lightweight DNN module to periodically predict weights across diverse architectures and datasets. WNN’s broader applicability enhanced the scope of weight forecasting to include tasks beyond basic image classification, making it a flexible booster for standard optimization methods. WNN was also applied to reverse forward training for machine unlearning \cite{jang2025learning}, demonstrating its feasibility in reversing learned processes. Recently, NiNo \cite{knyazev2024accelerating} advanced this concept further by incorporating inter-neuronal relationships into weight nowcasting, offering more nuanced predictions and additional training efficiency gains. 

\subsection{Learning-based Weight Initialization}
Effective weight initialization \cite{he2015delving,glorot2010understanding,zhu2021gradinit, yang2022towards} is crucial for training deep neural networks, as it significantly influences convergence speed and overall performance. Traditional methods \cite{he2015delving,glorot2010understanding} are based on statistical properties of the network layers. However, recent studies have introduced learning-based approaches that leverage meta-learning and hypernetworks to optimize initial weights.

MetaInit \cite{dauphin2019metainit}, a meta-learning algorithm was proposed to automate the search for optimal initializations. It operates on the hypothesis that suitable initializations facilitate gradient descent by positioning the optimization process in regions with minimal second-order effects, thereby enhancing training efficiency. 
Expanding on this concept, Knyazev et al. introduced a hypernetwork for weight initialization \cite{knyazev2024accelerating,knyazev2023can}. Using a pre-trained graph hypernetwork, the model predicts parameters for diverse unseen architectures, generating initial weights in a single forward pass.

\begin{figure*}[!t]
	\centering
    \begin{minipage}{0.85\linewidth}
	    \centering
	    {\includegraphics[trim={10 20 10 0},clip,width=0.95\columnwidth]{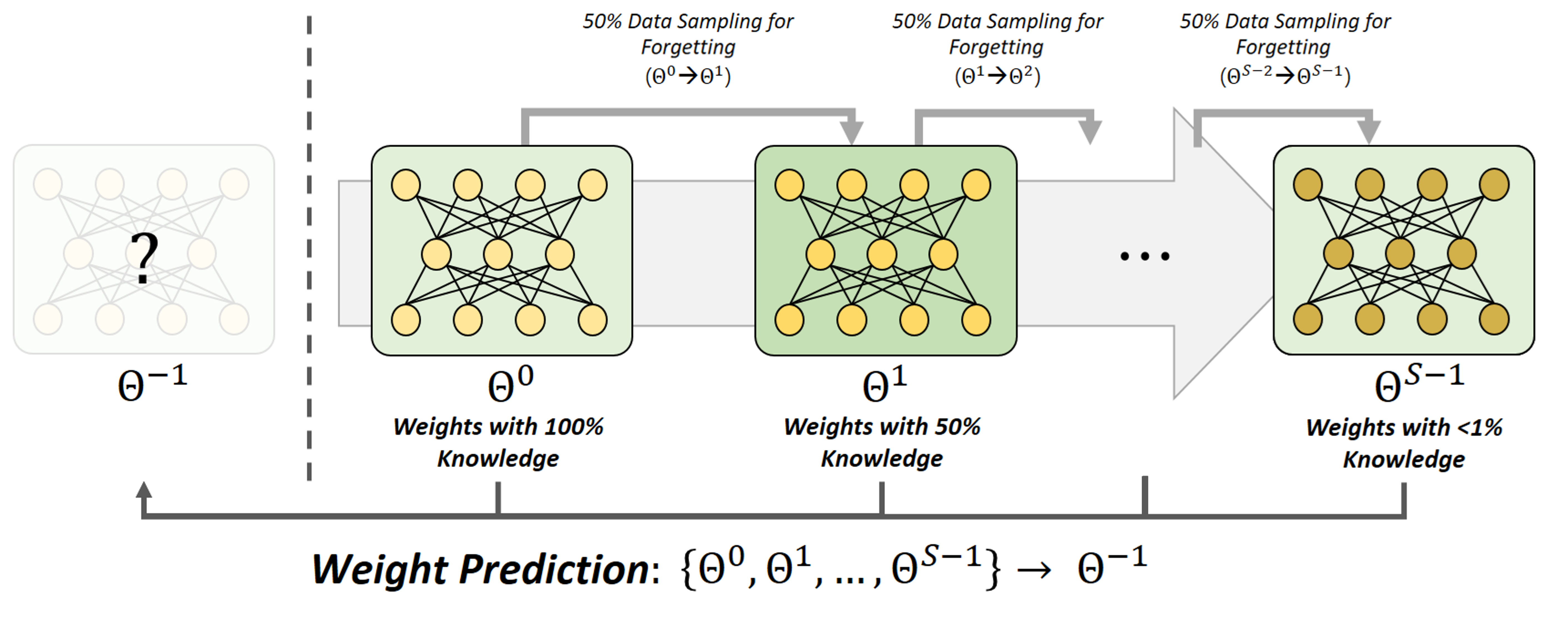}}   
    \end{minipage}%
	\caption{Conceptual Schematic of Knowledge Enrichment via Reversing the Progressive Forgetting.  }
	\label{fig:schematic}
    \vspace{-0.15cm}
\end{figure*}

\section{Problem Formulation}

\noindent\textbf{Motivation.} Fine-tuning is a fundamental technique in modern AI.
Yet, it often causes knowledge forgetting~\cite{kirkpatrick2017overcoming, chen2019catastrophic, luo2023empirical, shi2021overcoming}. This occurs when adapting to a subset of data overwrites knowledge about data outside that subset~\cite{mccloskey1989catastrophic, lopez2017gradient, goodfellow2013empirical}. Such forgetting has traditionally been regarded as a drawback of the training process.

Rather than treating forgetting as a drawback, we leverage it to obtain better pre-trained weights, inspired by prior studies as:

\begin{enumerate}
\item \textbf{Scaling Law in Dataset Size}: The more training data results in the better convergence, called scaling law \cite{kaplan2020scaling}.
\item \textbf{Forgetting via Fine-tuning}: Fine-tuning on the remaining data serves as a baseline unlearning method~\cite{golatkar2020eternal}.
\item \textbf{Fine-tuning Inversion}: Unlearning studies showed the reversibility of finetuning on specific subsets \cite{ilharco2022editing, jang2025learning}. 
\end{enumerate}

Based on these insights, we design a structured procedure that induces forgetting through sequential fine-tuning and then reverse it by leveraging weight transitions to restore lost knowledge. This inverse-forgetting process yields weights that retain richer information than the original pre-trained ones, effectively emulating the benefits of large-scale training without requiring additional data.

\begin{tcolorbox}[colback=gray!5!white, colframe=gray!75!black, title=Preliminary]
\vspace{-0.15cm}
\textit{Consider a dataset $D^0$ and a subset $D^1 \subset D^0$. Let $\Theta^0$ be the weights obtained from pre-training on $D^0$. When fine-tuning $\Theta^0$ on $D^1$, the resulting weights $\Theta^1$ exhibit a loss of knowledge about $D^0 \setminus D^1$ due to forgetting.}
\vspace{-0.15cm}
\end{tcolorbox}

\noindent We hypothesize that the relationship between $\Theta^1$ and $\Theta^0$ can be inverted to recover lost knowledge, as shown below:

\begin{tcolorbox}[colback=gray!5!white, colframe=gray!75!black, title=Hypothesis]
\vspace{-0.15cm}\textit{Given $\Theta^0$ trained on $D^0$, there exist ideal weights $\Theta^{-1}$ satisfying the following conditions:}
\begin{enumerate}
    \item $\Theta^{-1}$ is trained on $D^{-1}$, where $D^0\subset D^{-1}$.
    \item Fine-tuning $\Theta^{-1}$ on $D^0$ results in $\Theta^0$.
\end{enumerate}
\vspace{-0.15cm}
\end{tcolorbox}

\noindent This concept naturally extends in a progressive manner:

\begin{tcolorbox}[colback=gray!5!white, colframe=gray!75!black, title= Extension to Progressive Forgetting]
\vspace{-0.15cm}
\textit{Given a sequence of datasets $D^S\subset D^{S-1}\subset\dots\subset D^1\subset D^0$ constructed with a sampling ratio $r\in[0,1]$, we obtain a corresponding sequence of weights $[\Theta^S,\Theta^{S-1},\dots,\Theta^0]$, where each $\Theta^i$ is obtained by fine-tuning $\Theta^{i-1}$ on $D^i$.}
\vspace{-0.15cm}
\end{tcolorbox}
\noindent This formulation frames the problem as one of \textit{controlled forgetting}, where fine-tuning is systematically applied to progressively smaller datasets. Our goal is to analyze the resulting weight sequence and reverse the process to predict weights obtained from training on a larger dataset.

\section{Method}

Based on the problem formulation, we define our new framework, \textbf{Knowledge-Overflowed Weight (KNOW) Prediction}, illustrated in Fig.~\ref{fig1} and Fig.~\ref{fig:schematic}. It synthesizes knowledge-enriched weights that enhance training effectiveness without additional data. As an additional contribution, we also develop a meta-learned model named \textbf{Knowledge-Overflowed Weight Nowcaster (KNOWN)}, which is specifically designed for KNOW prediction.

\subsection{Knowledge-Overflowed Weight Prediction}
Once we obtain the sequence $[\Theta^S, \Theta^{S-1}, \dots, \Theta^0]$, our objective is to reverse this process. Specifically, we aim to predict weights $\Theta^{-1}$ that correspond to training on an ideal dataset $D^{-1}$, leveraging the successive fine-tuning trajectory as a basis. We define this process as \textit{KNOW Prediction}, which estimates weights corresponding to training on a larger dataset, thereby capturing enhanced knowledge. We model this via retrodiction of progressive forgetting as:
\begin{equation}
    Retrodiction:[(\Theta^{0}, \Theta^{1}, \Theta^{2}, \dots, \Theta^{S-1})] \rightarrow \Theta^{-1}.
\end{equation}
The predicted weights, denoted as $\hat{\Theta}^{-1} = Retrodiction([\Theta^0, \Theta^1, \Theta^2, \dots, \Theta^{S-1}])$, are referred to as \textit{KNOW}. Conceptually, this process is illustrated in Fig.~\ref{fig:schematic}. By reversing the intentional forgetting trajectory, KNOW prediction aims to recover weights that encode more knowledge than those trained solely on $D^0$. This aligns with our fundamental assumption: \textit{Pre-training on a larger dataset leads to better generalization performance}. 

\subsection{Transfer Learning to Downstream Task}

After predicting the weights with enhanced knowledge, we applied these predicted weights to transfer learning. As widely recognized, weights containing more comprehensive knowledge generally serve as a better initialization point for transfer learning. The convergence point on the new task and the resulting performance can then be used to validate the effectiveness of KNOW in extracting maximal knowledge from the given dataset.

\subsection{Knowledge-Overflowed Weight Nowcaster}
Prior work has explored the relationship between past and future weights to enhance DNN training efficiency. Building on the WNN \cite{jang2023learning}, we extend it by shifting to a dataset-size-based approach, allowing the model to reconstruct and even surpass the forgotten knowledge, approximating a state as if trained on more data.

Our \textbf{KNOWN} is a meta-trained hypernetwork that predicts weights trained on a larger dataset, capturing overcharged knowledge. This prediction is based on observed changes over an $S$-length sequence of weight updates:
\begin{align}
    &W^t_i = [\theta_{i}^{0},\theta_{i}^{1},\theta_{i}^{2},...,\theta_{i}^{S-1}],\\
    &dW^t_i = [\theta_{i}^{1}-\theta_{i}^{0}, ..., \theta_{i}^{S-1}-\theta_{i}^{S-2}],
\end{align}
where $i$ indexes the weight parameters and $N$ represents the total number of parameters in the target network. We set $S=5$ empirically. We then input $W^t_i$ and $dW^t_i$ into our KNOWN model to predict the residual toward the weight trained on a larger dataset:
\begin{align}
    &\hat{\theta}_{i}^{t-1}= \theta_{i}^{t} + KNOWN(W^t_i,dW^t_i),\\
    &\hat{\Theta} = \{ \hat{\theta}_{i} \}_{i=1,2,\dots,N}.
\end{align}

KNOWN follows the two-stream MLP architecture of WNN \cite{jang2023learning} and consists of 9,425 parameters. Further implementation details are provided in the Appendix.

\subsection{Meta Dataset Collection}

To facilitate meta-learning for our KNOWN, we constructed a dataset comprising weight trajectories with progressive forgetting, collected under diverse conditions, including variations in small architectures, datasets, sampling rates, and training strategies. Our data collection process involved continually sub-sampling the initial training dataset and progressively fine-tuning on the sampled subsets to induce forgetting of out-of-subset data. At the end of each fine-tuning phase, we stored the model weights, creating a structured dataset for meta-training. This process enables our model to learn from systematic weight evolution patterns and generalize across different training setups. Further details on the data collection are provided in Section \ref{sec:Collection}.

\subsection{Meta-training KNOWN}

Subsequently, we meta-trained our KNOWN with the objective of minimizing the $\ell_1$ residual error as :
\begin{equation}
\begin{split}
 & \| (\theta_{i}^{t}+KNOWN(W^t,dW^t)) - \theta_{i}^{t-1}  \|_{1}.    
\end{split}
 \label{eq:architecture_train}
\end{equation}
The $\ell_1$ benefits in high gradient for small values of $\theta$. Also, we categorized the DNN parameters into [convolution, fully-connected layer, bias] based on their corresponding operation, and created operation-specific KNOWN such as [$KNOWN_{Conv}$, $KNOWN_{FC}$, $KNOWN_{Bias}$].

\begin{figure}[!t]
	\centering
    \begin{minipage}{0.97\linewidth}
	    \centering
	    {\includegraphics[width=\columnwidth]{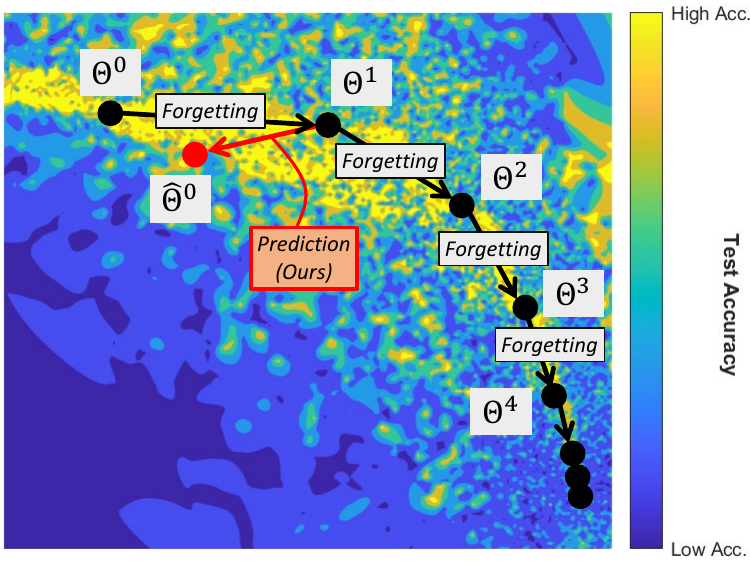}}   
    \end{minipage}%
    \vspace{-0.28cm}
	\caption{Landscape visualization of sequential forgetting and the weights with enriched knowledge prediction. }
	\label{fig:landscape}
    \vspace{-0.4cm}
\end{figure}

\subsection{Iterative Multi-step Forecasting}

In our proposed method, we utilize the sequence of weights $\left[\Theta^{0}, \Theta^{1}, \Theta^{2}, \dots, \Theta^{S-1}\right]$, which are obtained through progressive forgetting on the corresponding datasets $\left[D^{0}, D^{1}, D^{2}, \dots, D^{S-1}\right]$, to infer $\hat{\Theta}^{-1}$. If $\hat{\Theta}^{-1}$ proves to be sufficiently reliable, it becomes feasible to predict $\hat{\Theta}^{-2}$ using the sequence $\left[\hat{\Theta}^{-1}, \Theta^{0}, \Theta^{1}, \Theta^{2}, \dots, \Theta^{S-2}\right]$. Through this iterative approach, we can extract the maximum amount of knowledge from the available training datasets. The resulting $\hat{\Theta}$ can then serve as a well-initialized starting point, offering improved performance for fine-tuning on downstream tasks or enhancing results on the originally trained task.

\renewcommand{\thefootnote}{\fnsymbol{footnote}}
\begin{table*}[t]
    \renewcommand{\tabcolsep}{4.5mm}
	\centering
        \resizebox{0.99\linewidth}{!}
	{
        \begin{tabular}{>{\centering\arraybackslash}p{0.75cm}| c :c |ccccc} 
            \specialrule{.1em}{0em}{0em}
            \multirow{2}{*}[-0.em]{\bf\makecell{Pred.\\($\times$\textit{n})}} & \multicolumn{2}{c}{\multirow{2}{*}[-0.em]{\bf Methods}} & \multicolumn{5}{c}{\bf Amount of pre-training Data (\%)}\\\cdashline{4-8} 
            & \multicolumn{2}{c}{} &\bf 100\% & \bf 50\% & \bf 25\% & \bf 12.5\% & \bf 6.25\% \\\hline
            \multirow{2}{*}{\bf$\times$1}&\multicolumn{2}{c|}{Na\"ive Transfer (Baseline)} & $92.40\pm0.11$ & $92.08\pm0.18$ & $91.90\pm0.24$ & $91.49\pm0.22$ & $91.51\pm0.12$ \\
            &\multicolumn{2}{c|}{Incremental Pretraining (Baseline)} & $92.29\pm0.10$ & $91.32\pm0.14$ & $90.79\pm0.48$ & $90.07\pm0.10$ & $89.51\pm0.08$ \\\hline
            \multirow{8}{*}[-0.3em]{\bf$\times$2}& 
            {\multirow{8}{*}{\bf KNOW\footnotemark[4]}} & LinearFit & $92.70\pm0.16$ & $92.28\pm0.16$ & $91.89\pm0.24$& $91.42\pm0.15$ & $91.37\pm0.16$ \\
            &  & LogFit & \underline{$92.83\pm0.25$} & \underline{$92.41\pm0.07$}& $91.89\pm0.28$&  $91.39\pm0.17$& $91.33\pm0.24$ \\
            &  & ExpFit & $79.58\pm0.41$ & $86.79\pm0.23$& $89.26\pm0.04$& $89.85\pm0.08$ & $90.60\pm0.16$\\
            &  & TaskVector  \cite{ilharco2022editing} & $92.69\pm0.09$ & $92.39\pm0.12$ & \underline{$92.22\pm0.25$} & $91.45\pm0.05$ &  $91.46\pm0.08$\\
            &  & ConsensusTA\footnotemark[2]  \cite{wang2024localizing}& $92.69\pm0.09$ & $92.39\pm0.12$ & $92.22\pm0.25$ & $91.45\pm0.05$ &  $91.46\pm0.08$\\     
            &  & MagMax\footnotemark[3] \cite{marczak2024magmax} & $92.69\pm0.09$ & $92.39\pm0.12$ & $92.22\pm0.25$ & $91.45\pm0.05$ &  $91.46\pm0.08$\\
            &  & TSV  \cite{gargiulo2025task} & $92.82\pm0.14$ & $92.36\pm0.16$  &  $92.19\pm0.11$& \underline{$91.64\pm0.19$} &  \underline{$91.74\pm0.25$}\\
            \cdashline{3-8}\rule{0pt}{10.2pt} 
            &  & \cellcolor{gray!20}\bf KNOWN & \cellcolor{gray!20}\bf 93.00 $\pm$ 0.11 & \cellcolor{gray!20}\bf 92.58 $\pm$ 0.14 & \cellcolor{gray!20}\bf 92.29 $\pm$ 0.04 & \cellcolor{gray!20}\bf 92.11 $\pm$ 0.10 & \cellcolor{gray!20}\bf 91.90 $\pm$ 0.13 \\\hline
            
            \multirow{8}{*}[-0.3em]{\bf$\times$4}& 
            \multirow{8}{*}{\bf KNOW\footnotemark[4]} & LinearFit & $92.40\pm0.08$ & $91.99\pm0.22$ & $91.63\pm0.14$&  $90.52\pm0.12$& $90.64\pm0.09$ \\
            &  & LogFit & \underline{$92.82\pm0.10$} & $92.33\pm0.13$ & $91.87\pm0.13$&  $91.05\pm0.15$& $91.08\pm0.23$ \\
            &  & ExpFit & $72.70\pm0.64$ &  $80.75\pm0.55$ & $85.16\pm0.04$&  $87.46\pm0.27$& $88.80\pm0.29$\\
            &  & TaskVector  \cite{ilharco2022editing} & $92.70\pm0.09$ & $92.48\pm0.10$ & $92.19\pm0.12$ & $91.45\pm0.14$ & $91.36\pm0.18$ \\
                &  & ConsensusTA\footnotemark[3]  \cite{wang2024localizing} & $92.70\pm0.09$ & $92.48\pm0.10$ & $92.19\pm0.12$ & $91.45\pm0.14$ & $91.36\pm0.18$ \\
            &  & MagMax  \cite{marczak2024magmax} & $92.44\pm0.28$ & $92.19\pm0.22$ & $92.12\pm0.19$ &  $91.29\pm0.16$& $91.15\pm0.14$ \\
            &  & TSV  \cite{gargiulo2025task} & $92.74\pm0.13$ & \underline{$92.40\pm0.16$} &  \underline{$92.17\pm0.18$}&  \underline{$91.64\pm0.20$} & \underline{$91.63\pm0.16$} \\
            \cdashline{3-8}
            \rule{0pt}{10.2pt} 
            &  & \cellcolor{gray!20} \bf KNOWN & \cellcolor{gray!20}\bf 93.27 $\pm$ 0.09 & \cellcolor{gray!20}\bf 92.62 $\pm$ 0.25 & \cellcolor{gray!20}\bf 92.88 $\pm$ 0.11 & \cellcolor{gray!20}\bf 92.40 $\pm$ 0.06 & \cellcolor{gray!20}\bf 91.98 $\pm$ 0.14 \\\hline
            
            \multirow{8}{*}[-0.3em]{\bf$\times$8}& 
            \multirow{8}{*}{\bf KNOW\footnotemark[4]} & LinearFit & $92.13\pm0.10$ & $91.71\pm0.22$ & $90.77\pm0.18$&  $90.15\pm0.40$& $90.42\pm0.44$  \\
            &  & LogFit & $92.65\pm0.10$ & $91.93\pm0.25$& $91.59\pm0.05$ &  $90.49\pm0.25$& $90.68\pm0.24$\\
            &  & ExpFit & $66.54\pm0.29$ &$73.89\pm0.91$ & $81.05\pm0.14$ &  $84.34\pm0.56$& $85.16\pm0.45$ \\
            &  & TaskVector  \cite{ilharco2022editing} & $92.65\pm0.13$ & $92.24\pm0.15$ & $92.03\pm0.15$ & $91.13\pm0.21$ & $91.28\pm0.34$ \\
            &  & ConsensusTA  \cite{wang2024localizing} & $92.43\pm0.14$ & $92.02\pm0.12$ & $91.93\pm0.17$ & $91.11\pm0.29$ & $90.63\pm0.32$ \\
            &  & MagMax  \cite{marczak2024magmax} & $92.39\pm0.20$ & $92.16\pm0.10$ & $92.07\pm0.09$ &  $91.43\pm0.20$& $90.85\pm0.33$ \\
            &  & TSV  \cite{gargiulo2025task} & \underline{$92.72\pm0.11$} &  \underline{$92.37\pm0.15$}& \underline{$92.25\pm0.15$} & \underline{$91.51\pm0.17$} &  \underline{$91.48\pm0.15$}\\
            \cdashline{3-8}
            \rule{0pt}{10.2pt} 
            &  & \cellcolor{gray!20} \bf KNOWN & \cellcolor{gray!20}\bf 93.55 $\pm$ 0.05 & \cellcolor{gray!20}\bf 93.11 $\pm$ 0.19 & \cellcolor{gray!20}\bf 92.92 $\pm$ 0.15 & \cellcolor{gray!20}\bf 92.22 $\pm$ 0.37 & \cellcolor{gray!20}\bf 92.07 $\pm$ 0.18 \\
            \specialrule{.1em}{.0em}{-.1em}
            \end{tabular}
        }
        \vspace{-0.25cm}
        \caption{Experimental results for ResNet18 with CIFAR100 (pre-training) and CIFAR10 (Downstream). Each value represents the test accuracy of CIFAR10. §Note that all results, except \textit{Naïve Transfer}, utilize \textbf{the proposed KNOW prediction} with various methods. }
        \vspace{-0.3cm}
    \label{table:t1}
\end{table*}

\subsection{Qualitative Validation of the Proposed Concept}
To validate our problem definition, we first visualized the trajectory of the sequence 
\( \left[\Theta^{0}, \Theta^{1}, \Theta^{2}, \dots, \Theta^{S}\right] \), 
obtained through sequential forgetting, alongside the surrounding loss landscape. We then predicted KNOW \( \hat{\Theta}^{0} \) using 
\( \left[\Theta^{1}, \Theta^{2}, \dots, \Theta^{S}\right] \) 
and compared it to the true weight \( \Theta^{0} \). For precise visualization, we trained a small-scale vanilla CNN (\(\approx 25K\) parameters) on the CIFAR10 dataset \cite{cifar10} and stored its weight trajectory by sequentially downsampling the dataset and fine-tuning. Over 100K points and their corresponding test accuracies were recorded through exhaustive noise injection along the trajectory. Additionally, we predicted the weight \( \hat{\Theta}^{0} \) using our KNOWN.  

To reduce dimensionality for visualization, we applied Principal Component Analysis (PCA) on the collected weights. The first two principal components and the associated test accuracies were used to construct a loss landscape, with the weight trajectory mapped onto this landscape, as shown in Fig. \ref{fig:landscape}. Notably, the sequence  
\( \left[\Theta^{0}, \Theta^{1}, \Theta^{2}, \dots, \Theta^{S}\right] \)  
forms a smooth curve, with regions around the curve exhibiting high test accuracy. This visualization provides two key insights: 1) Forgetting through sequential downsampling leads to gradual, rather than abrupt, changes in the weight space, and 2) There exists a pathway connecting the converged weights, consistent with findings from Mode Connectivity studies \cite{draxler2018essentially, garipov2018loss, benton2021loss}.  

Notably, our predicted weight \( \hat{\Theta}^{0} \) is positioned closer to the true weight \( \Theta^{0} \) than to \( \Theta^{1} \), as illustrated in Fig. \ref{fig:landscape}. This empirically supports the feasibility of our approach, demonstrating that \( \hat{\Theta}^{0} \) more accurately approximates \( \Theta^{0} \) than \( \Theta^{1} \). Also, the left side of \( \Theta^{0} \) exhibits higher test accuracy, reinforcing the potential of recurrent weight prediction.

\begin{table*}[t]
    \renewcommand{\tabcolsep}{1.0mm}
	\centering
        \resizebox{0.99\linewidth}{!}
	{
        \begin{tabular}{>{\centering\arraybackslash}p{0.8cm}| c:c | ccccc} 
            \specialrule{.1em}{0em}{0em}
            \multirow{2}{*}[-0.em]{\bf\makecell{Pred.\\($\times$\textit{n})}} & \multicolumn{2}{c|}{\multirow{2}{*}[-0.1em]{\bf Methods}} & \multicolumn{5}{c}{\bf Downstream Dataset $\mathcal{D}$ (ImageNet (Pre-train) $\rightarrow$ $\mathcal{D}$)}\\\cline{4-8} 
            & \multicolumn{2}{c|}{} &\bf CIFAR100 & \bf TinyImageNet & \bf Car & \bf CUB & \bf Flowers \\\hline
            \rule{0pt}{10.0pt}\bf$\times$1&\multicolumn{2}{c|}{Na\"ive Transfer (Baseline)} & $82.03\pm0.25$& $76.17\pm0.33$& $88.12\pm0.35$& $70.49\pm0.58$& $87.98\pm0.38$\\\hline
            \multirow{4}{*}[-0.3em]{\bf$\times$3}
            & {\multirow{4}{*}{\bf KNOW}} & LogFit & $81.73\pm0.37$ ($-0.30$) & $77.32\pm0.14$ ($+1.15$)& $88.49\pm0.68$ ($+0.37$)& $70.45\pm0.71$ ($-0.04$)& \underline{$88.24\pm0.20$} ($+0.26$)\\
            &  & TaskVector  \cite{ilharco2022editing} & \underline{$82.15\pm0.29$} ($+0.12$)& \underline{$77.49\pm0.21$} ($+1.32$)& $88.31\pm0.24$ ($+0.19$)& \underline{71.00 $\pm$ 0.25} ($+0.51$)& $88.18\pm0.42$ ($+0.20$)\\
            &  & TSV  \cite{gargiulo2025task} & $82.14\pm0.46$ ($+0.11$)& $76.06\pm0.17$ ($-0.11$) & \bf 88.78 $\pm$ 0.12 ($+0.80$) & $70.84\pm0.48$ ($+0.35$) & $86.37\pm0.37$ ($-1.61$)\\
            \cdashline{3-8}\rule{0pt}{10.2pt} 
            &  & \cellcolor{gray!20}\bf KNOWN & \cellcolor{gray!20}\bf 82.46 $\pm$ 0.20 ($+0.43$)& \cellcolor{gray!20}\bf 77.53 $\pm$ 0.11 ($+1.36$)& \cellcolor{gray!20}\underline{88.57 $\pm$ 0.18} ($+0.45$)& \cellcolor{gray!20}\bf{71.18 $\pm$ 0.45} ($+0.69$)& \cellcolor{gray!20}\bf 88.65 $\pm$ 0.69 ($+0.67$)\\\hline
            
            \multirow{5}{*}[-0.3em]{\bf$\times$9}& 
            {\multirow{5}{*}{\bf KNOW}} & LogFit & $81.75\pm0.35$ ($-0.28$)& \underline{$77.29\pm0.14$} ($+1.12$)& $88.42\pm0.30$ ($+0.30$)& \underline{$71.09\pm0.65$} ($+0.60$)& \underline{$88.43\pm0.57$} ($+0.45$)\\
            &  & TaskVector  \cite{ilharco2022editing} & $82.12\pm0.45$ ($+0.09$)& \underline{$77.29\pm0.23$} ($+1.12$)& $88.18\pm0.17$ ($+0.06$)& $70.38\pm0.34$ ($-0.11$)& $88.37\pm0.43$ ($+0.39$)\\
            &  & MagMax  \cite{marczak2024magmax} & $82.27\pm0.24$ ($+0.24$)& $75.98\pm0.18$ ($-0.19$) & $88.61 \pm 0.31$ ($+0.49$)& $70.95\pm0.39$ ($+0.46$)& $86.51\pm0.76$ ($-1.47$)\\
            &  & TSV  \cite{gargiulo2025task} & \underline{$82.31 \pm 0.20$} ($+0.28$)& $76.18\pm0.27$ ($+0.01$) & \bf 88.85 $\pm$ 0.09 ($+0.73$) & $70.79\pm0.32$ ($+0.30$)& $86.47\pm0.56$ ($-1.51$)\\
            \cdashline{3-8}\rule{0pt}{10.2pt} 
            &  & \cellcolor{gray!20}\bf KNOWN & \cellcolor{gray!20}\bf 82.33 $\pm$ 0.15 ($+0.30$)& \cellcolor{gray!20}\bf 77.58 $\pm$ 0.12 ($+1.41$)& \cellcolor{gray!20}\bf 88.85 $\pm$ 0.23 ($+0.73$)& \cellcolor{gray!20}\bf 71.30 $\pm$ 0.28 ($+0.81$)& \cellcolor{gray!20}\bf 88.53 $\pm$ 0.27 ($+0.55$)\\
            \specialrule{.1em}{.0em}{-.1em}
            \end{tabular}
        }
        \vspace{-0.25cm}
        \caption{Applications of KNOW prediction for ImageNet (Pre-training) and various image classification datasets (Downstream). Each value represents the test accuracy of each downstream dataset. }
        \vspace{-0.1cm}
    \label{table:t2}
\end{table*}

\begin{table*}[t]
    \renewcommand{\tabcolsep}{1.0mm}
	\centering
        \resizebox{0.99\linewidth}{!}
	{
        \begin{tabular}{>{\centering\arraybackslash}p{0.8cm}| c:c | cccc|c} 
            \specialrule{.1em}{0em}{0em}
            \multirow{3}{*}[-0.em]{\bf\makecell{Pred.\\($\times$\textit{n})}} & \multicolumn{2}{c|}{\multirow{3}{*}[-0.3em]{\bf Methods}} & \multicolumn{4}{c|}{\bf Leave-One-Domain-Out} & \multirow{3}{*}[-0.em]{\bf Avg.}\\\cdashline{4-7} 
             &  \multicolumn{2}{|c|}{} &\textbf{[sketch, cartoon, photo]} & \textbf{[art, cartoon, photo]} & \textbf{[sketch, art, photo]} & \textbf{[sketch, art, cartoon]} &\multirow{3}{*}[-0.em]{}\\
            & \multicolumn{2}{|c|}{} &\textbf{$\rightarrow$ art} & \textbf{$\rightarrow$ sketch} & \textbf{$\rightarrow$ cartoon} & \textbf{$\rightarrow$ photo} &\\\hline
            \rule{0pt}{10.0pt}\bf$\times$1&\multicolumn{2}{c|}{Na\"ive Transfer (Baseline)} & $66.36\pm1.04$ & $42.12\pm0.14$ & $54.65\pm1.36$ & $90.78\pm0.42$ &  63.48 \\\hline
            \multirow{4}{*}[-0.3em]{\bf$\times$3}
            & {\multirow{4}{*}{\bf KNOW}} & LogFit & $67.49\pm0.61$ ($+1.13$)& \bf 48.58 $\pm$ 1.66 ($+6.46$)& $60.73\pm0.33$ ($+6.08$)& $86.16\pm0.78$ ($-4.62$)&  65.74 ($+2.26$)\\
            &  & TaskVector  \cite{ilharco2022editing} & \underline{$69.04\pm0.32$} ($+2.68$)& $43.85\pm0.69$ ($+1.73$)& \underline{$60.83\pm0.60$} ($+6.18$)& \underline{$92.53\pm0.41$} ($+1.75$)& \underline{66.56} ($+3.08$)\\
            &  & TSV  \cite{gargiulo2025task} & $65.50\pm0.87$ ($-0.86$)& $41.73\pm1.35$ ($-0.39$) & $55.11\pm0.83$ ($+0.46$)& \bf  $89.56\pm0.86$ ($-1.22$)& 62.97 ($-0.51$)\\
            \cdashline{3-8}\rule{0pt}{10.2pt} 
            &  & \cellcolor{gray!20}\bf KNOWN & \cellcolor{gray!20}\bf 72.12 $\pm$ 0.27 ($+5.76$)& \cellcolor{gray!20}\underline{44.11 $\pm$ 1.52} ($+1.99$)& \cellcolor{gray!20}\bf 62.73 $\pm$ 1.27 ($+8.08$)& \cellcolor{gray!20}\bf 93.87 $\pm$ 1.10 ($+3.09$)& \cellcolor{gray!20}\bf 68.21 ($+4.73$)\\\hline
            
            \multirow{5}{*}[-0.3em]{\bf$\times$9}& 
            {\multirow{5}{*}{\bf KNOW}} & LogFit & \underline{$69.01\pm0.38$} ($+2.65$)& \bf 44.69 $\pm$ 1.01 ($+2.57$)& $60.45\pm0.33$ ($+5.80$)& $91.97\pm0.60$ ($+1.19$)& 66.53 ($+3.05$)\\
            &  & TaskVector  \cite{ilharco2022editing} & $68.89\pm0.91$ ($+2.53$)& $43.43\pm0.95$ ($+1.31$)& \underline{$60.89\pm0.74$} ($+6.24$)& \bf  93.08 $\pm$ 0.20 ($+2.30$)& \underline{66.57} ($+3.09$)\\
            &  & MagMax  \cite{marczak2024magmax} & $64.77\pm0.68$ ($-1.59$)& $41.78\pm1.32$ ($-0.34$)& $55.76\pm0.88$ ($+1.11$)& \bf  $90.37\pm0.78$ ($-0.41$)& 63.17 ($-0.31$)\\
            &  & TSV  \cite{gargiulo2025task} & $65.27\pm1.13$ ($-1.09$)& $43.07\pm0.98$ ($+0.95$)& $55.36\pm0.69$ ($+0.71$)& \bf  $90.10\pm0.71$ ($-0.68$)& 63.45 ($-0.03$)\\
            \cdashline{3-8}\rule{0pt}{10.2pt} 
            &  & \cellcolor{gray!20}\bf KNOWN & \cellcolor{gray!20}\bf 72.07 $\pm$ 0.20 ($+5.71$)& \cellcolor{gray!20}\underline{44.02 $\pm$ 0.78} ($+1.90$)& \cellcolor{gray!20}\bf 64.28 $\pm$ 0.88 ($+9.63$)& \cellcolor{gray!20}\underline{92.98 $\pm$ 0.29} ($+2.20$)& \cellcolor{gray!20}\bf 68.33 ($+4.85$)\\
            \specialrule{.1em}{.0em}{-.1em}
            \end{tabular}
        }
        \vspace{-0.2cm}
        \caption{Results about domain generalization using PACS dataset. Each value indicates the accuracy on the domain right sided of "$\rightarrow$"}
        \vspace{-0.3cm}
    \label{table:4}
\end{table*}

\footnotetext[2]{ConsensusTA uses majority voting and is equivalent to the Task Vector when no majority exists (fewer than three candidates: ×2 or ×4)}
\footnotetext[3]{MagMax selects coordinate-wise maximum updates among candidates, making it equivalent to Task Vector in the single-candidate case (×2).}
\section{Experiments}
This section presents experiments demonstrating the feasibility of our KNOW prediction and KNOWN. Note that we used the meta-trained KNOWN for all experiments \textbf{without extra meta-data collection or meta-training.} All results (except Naïve Transfer) employ the proposed KNOW prediction through various methods, and \textbf{the gains should be evaluated relative to the Naïve Transfer.}

\subsection{Training Data Collection for KNOWN} \label{sec:Collection}
Our approach incorporates an hyper-model, \textbf{KNOWN}, designed to predict weights with augmented knowledge. This requires generating weight trajectories through sequential fine-tuning with intentional forgetting. To train KNOWN, we collected weight trajectories from multiple small-scale DNNs (e.g., CNN, ResNet \cite{resnet}, DenseNet \cite{densenet}, ShuffleNet \cite{shufflenetv2}, MobileNetV2 \cite{sandler2018mobilenetv2}), each with fewer than 3M parameters. Using CIFAR10 \cite{cifar10}, MNIST \cite{lecun1998gradient}, and Fashion MNIST \cite{fashionmnist}, we applied random sampling at each fine-tuning stage to vary data exposure and induce forgetting. For each trial, we randomly selected a sampling rate \( r \), learning rate, and batch size. The full dataset \( D^0 \) was progressively subsampled as \( [D^{S-1} \subset \dots \subset D^1 \subset D^0] \), with each \( D^{i+1} \) obtained by sampling from \( D^i \) according to \( r \). Starting from base weights \( \Theta^0 \) trained on \( D^0 \), we iteratively finetuned on smaller subsets, capturing the effects of forgetting and generating weight trajectories for KNOW prediction. This process produced \(\sim 50\)GB of trajectory data, which was then used to train KNOWN with the objective in Eq. \eqref{eq:architecture_train}. Once trained, KNOWN generalizes across all settings in this study without additional training cost, making it an efficient predictor of enriched knowledge.

\subsection{Results of Image Classification Task} \label{sec:resnet}

The proposed method predicts synthetic weights with enriched knowledge, serving as effective initializations for transfer learning. Greater pre-trained knowledge is known to accelerate convergence and improve downstream performance.  
To verify this, we first trained ResNet18 \cite{resnet} on CIFAR100 \cite{cifar10} for 200 epochs using the Adam optimizer with cosine decay and data augmentation. We then sequentially finetuned it on progressively smaller subsets of the training dataset with \( r = 0.5 \), producing the weight trajectory  
\( \left[\Theta^{0}, \Theta^{1}, \Theta^{2}, \dots, \Theta^{S-1}\right] \).  
Next, we iteratively predicted a set of KNOW weights  
\( [\hat{\Theta}^{-1},\hat{\Theta}^{-2}, \hat{\Theta}^{-3}] \),  
denoted as \( \times2 \), \( \times4 \), and \( \times8 \). Note that we used \( r = 0.5 \), meaning the predicted weights are denoted as \( \times2 \), \( \times4 \), and \( \times8 \). These enriched weights were then finetuned on the CIFAR10 dataset. Additionally, we repeated this process starting from  
[100\%, 50\%, 25\%, 12.5\%, 6.25\%] of the CIFAR100 training set to assess robustness to dataset size.  

For comparison, we implemented the na\"ive transfer, incremental pretraining and regression-based methods for KNOW prediction, including curve fitting approaches (i.e., Linear, Log, and Exponential functions), Task Vector \cite{ilharco2022editing}, MagMax \cite{marczak2024magmax}, ConsensusTA \cite{wang2024localizing}, and TSV \cite{gargiulo2025task}. Specifically, Na\"ive transfer represents the fine-tuning case starting from \( \Theta^{0} \) without KNOW prediction. Incremental pre-training involves progressive training from 6.25\% to 100\% of the data, followed by a transfer without KNOW prediction. 
For KNOW prediction, curve fitting extrapolates the weight trajectory \( [\Theta^{0},\Theta^{1},\dots,\Theta^{S-1}] \) to \([-1,-2,-3]\) in a coordinate-wise manner. Task Vector predicts through linear extrapolation as $\hat{\Theta}^{-s}=\Theta^{0}+\lambda(\Theta^{0}-\Theta^{s})$ for $s=1,2,...,S$ with \( \lambda=0.2 \). 
MagMax extends Task Vector by selecting, for each coordinate, the update with the largest absolute change (reducing to Task Vector when \(S=2\)). ConsensusTA keeps only parameters updated in the majority of task vectors, thus filtering task-specific noise, and was applicable only to the \(\times 8\) case for majority voting. TSV applies SVD to per-layer task matrices and retains the top 10\% components to stabilize the update.
All methods followed the same training protocol with ten repetitions.

As shown in Table \ref{table:t1}, incremental pretraining underperforms due to the overfitted initialization problem, as detailed in the Supplementary. In contrast, KNOWN consistently enhance performance, whereas other methods sometimes fail. This highlights the difficulty of modeling the forgetting trajectory with fixed curve functions, as our landscape visualization in Fig. \ref{fig:landscape} shows its complex, non-linear nature. When the curve model diverges from the true trajectory, performance drops sharply, as in ExpFit. For our approach, the results demonstrate that the proposed method successfully enriches the knowledge embedded in the weights, leading to improved CIFAR10 performance after convergence. Notably, the KNOW outperform a baseline trained on a larger dataset, confirming that our approach effectively predicts weight configurations with enhanced knowledge, beneficial for fine-tuning. Interestingly, the iterative predictions further improve fine-tuning performance, indicating that our predicted weights are not only reliable but also reusable for subsequent predictions.

We validated the generality of our method on a complex DNN and diverse datasets without additional fine-tuning of KNOWN. Specifically, we used PVTv2 \cite{wang2021pvtv2}, pre-trained on ImageNet \cite{deng2009imagenet}. After sequential fine-tuning with \( r=0.33 \), we synthesized \( [\hat{\Theta}^{-1}, \hat{\Theta}^{-2}] \) and applied them to CIFAR100, TinyImageNet \cite{tiny_imagenet_challenge}, Stanford Cars \cite{KrauseStarkDengFei}, CUB \cite{wah2011caltech}, and Oxford Flowers \cite{Nilsback08}, all with over 100 classes. With \( r=0.33 \), the knowledge scaling factor is expected to be \( \times3 \) and \( \times9 \). As shown in Table \ref{table:t2}, our method consistently improves performance at convergence, whereas alternatives like LogFit and Task Vector often degrade or yield only marginal gains. These results highlight the broad applicability and robustness of our approach across datasets and values of \( r \).

\begin{table}[t]
    \renewcommand{\tabcolsep}{3mm}
	\centering
        \resizebox{\linewidth}{!}
	{
        \begin{tabular}{c | c:c | c} 
            \specialrule{.1em}{0em}{0em}
            \bf Pred.($\times$\textit{n}) & \multicolumn{2}{c|}{\multirow{1}{*}{\bf Methods}} & \bf Accuracy (\%)\\\hline
            \rule{0pt}{10.0pt}\bf$\times$1&\multicolumn{2}{c|}{Na\"ive Transfer (Baseline)} & $37.14\pm0.21$ \\\hline

            \multirow{4}{*}[-0.3em]{\bf$\times$3}& 
            {\multirow{4}{*}{\bf KNOW}} & LogFit & \underline{$39.21\pm0.28$} ($+2.07$)\\
            &  & TaskVector  \cite{ilharco2022editing} & $39.12\pm0.28$ ($+1.98$)\\
            &  & TSV  \cite{gargiulo2025task} & $37.32\pm0.21$ ($+0.18$)\\
            \cdashline{3-4}\rule{0pt}{10.2pt} 
            &  & \cellcolor{gray!20}\bf KNOWN &\cellcolor{gray!20}\bf 39.38 $\pm$ 0.15 ($+2.24$)\\\hline

            \multirow{5}{*}[-0.3em]{\bf$\times$9}& 
            {\multirow{5}{*}{\bf KNOW}} & LogFit & $39.11\pm0.19$ ($+1.97$)\\
            &  & TaskVector  \cite{ilharco2022editing} & \underline{$39.20\pm0.20$} ($+2.06$)\\
            &  & MagMax  \cite{marczak2024magmax} & $37.20\pm0.22$ ($+0.06$)\\
            &  & TSV  \cite{gargiulo2025task} & $37.15\pm0.17$ ($+0.01$)\\
            \cdashline{3-4}\rule{0pt}{10.2pt} 
            &  & \cellcolor{gray!20}\bf KNOWN &\cellcolor{gray!20}\bf39.25 $\pm$ 0.17 ($+2.11$)\\
            \specialrule{.1em}{.0em}{-.1em}
            \end{tabular}
        }
        \vspace{-0.2cm}
        \caption{Application of knowledge-overflowed PVTv2 to Flickr8K image captioning (Masked Accuracy).}
        \vspace{-0.1cm}
\label{table:3}
\end{table}

\begin{table}[t]
    \renewcommand{\tabcolsep}{3mm}
	\centering
        \resizebox{\linewidth}{!}
	{
        \begin{tabular}{c | c:c | c} 
            \specialrule{.1em}{0em}{0em}
            \bf Pred.($\times$\textit{n}) & \multicolumn{2}{c|}{\multirow{1}{*}{\bf Methods}} & \bf  mIoU (\%)\\\hline
            \rule{0pt}{10.0pt}\bf$\times$1&\multicolumn{2}{c|}{Na\"ive Transfer (Baseline)} & $68.52 \pm 1.34$ \\\hline

            \multirow{4}{*}[-0.3em]{\bf$\times$3}& 
            {\multirow{4}{*}{\bf KNOW}} & LogFit & $68.79 \pm 0.68$ ($+0.27$)\\
            & & TaskVector  \cite{ilharco2022editing} & \underline{$68.83 \pm 1.71$} ($+0.31$)\\
            & & TSV  \cite{gargiulo2025task} & $68.64 \pm 0.71$ ($+0.12$)\\
            \cdashline{3-4}\rule{0pt}{10.2pt} 
            &  &\cellcolor{gray!20}\bf  KNOWN &\cellcolor{gray!20}\bf 69.00 $\pm$ 1.04 ($+0.48$)\\\hline

            \multirow{5}{*}[-0.3em]{\bf$\times$9}& 
            {\multirow{5}{*}{\bf KNOW}} & LogFit & $69.16\pm1.95$ ($+0.64$)\\
            &  & TaskVector  \cite{ilharco2022editing} & $67.99\pm0.96$ ($-0.53$)\\
            & & MagMax  \cite{marczak2024magmax} & \underline{$70.04\pm0.81$} ($+1.52$)\\
            & & TSV  \cite{gargiulo2025task} & $68.98 \pm 1.06$ ($+0.46$)\\
            \cdashline{3-4}\rule{0pt}{10.2pt} 
            & &\cellcolor{gray!20}\bf KNOWN &\cellcolor{gray!20}\bf71.22 $\pm$ 0.82 ($+2.70$)\\
            \specialrule{.1em}{.0em}{-.1em}
            \end{tabular}
        }
        \vspace{-0.2cm}
        \caption{Application of KNOW prediction to DeepLabV3+ with Cityscapses image segmentation (mIoU).}
        \vspace{-0.3cm}
    \label{table:5}
\end{table}

\begin{figure}[!t]
	\centering
    \begin{minipage}{0.25\linewidth}
	    \centering
	    {\includegraphics[width=\columnwidth]{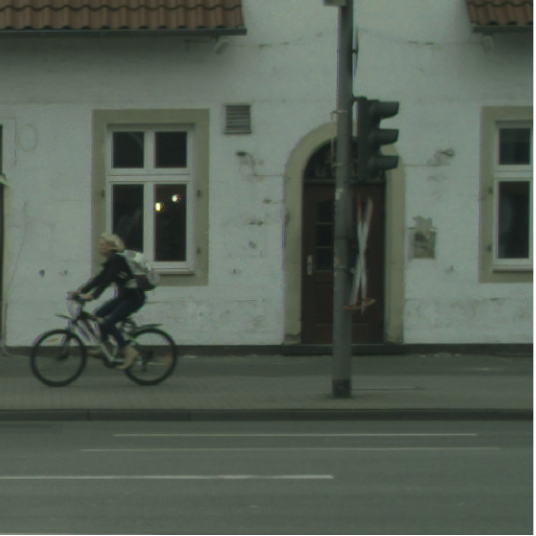}} 
    \end{minipage}%
    \begin{minipage}{0.25\linewidth}
	    \centering
	    {\includegraphics[width=\columnwidth]{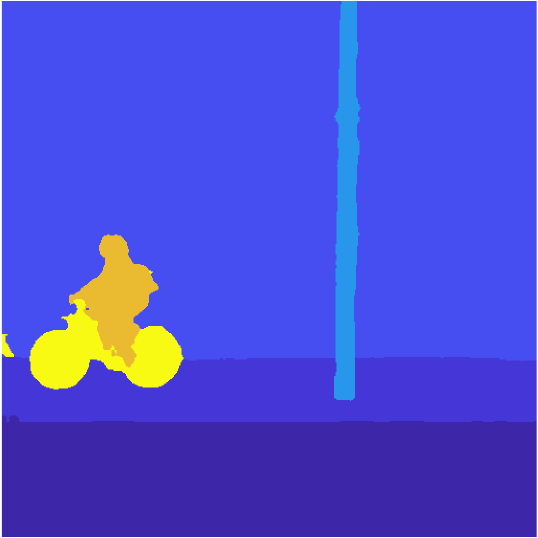}}   
    \end{minipage}%
    \begin{minipage}{0.25\linewidth}
	    \centering
	    {\includegraphics[width=\columnwidth]{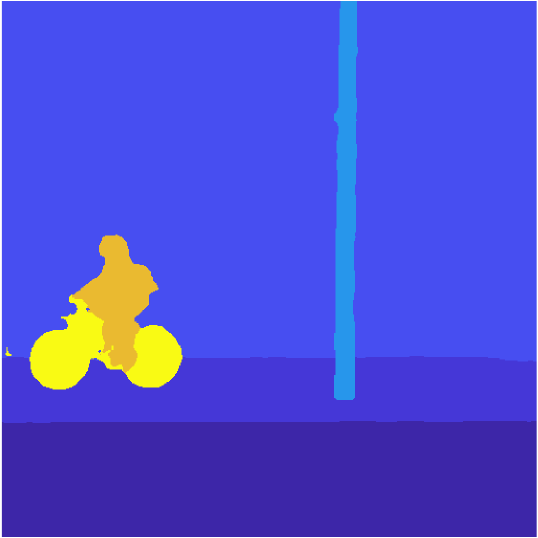}}   
    \end{minipage}%
    \begin{minipage}{0.25\linewidth}
	    \centering
	    {\includegraphics[width=\columnwidth]{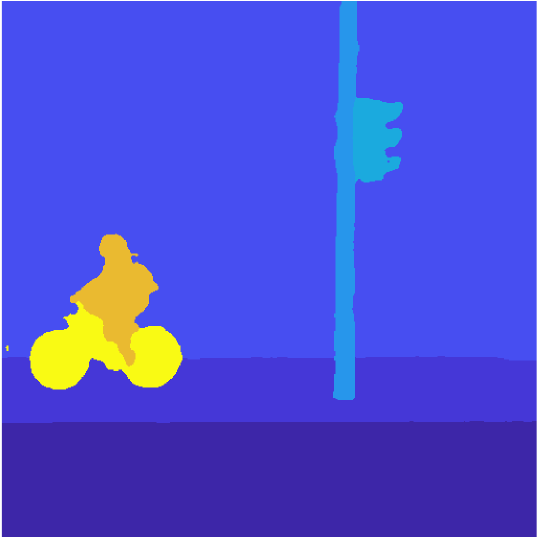}}   
    \end{minipage}%
    
    \begin{minipage}{0.25\linewidth}
	    \centering
	    {\includegraphics[width=\columnwidth]{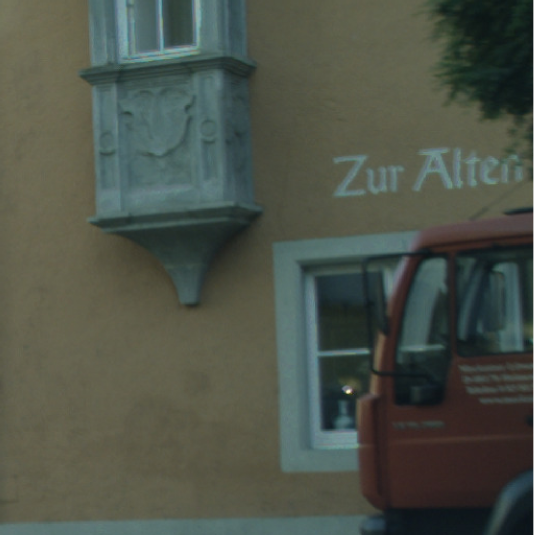}} 
    \end{minipage}%
    \begin{minipage}{0.25\linewidth}
	    \centering
	    {\includegraphics[width=\columnwidth]{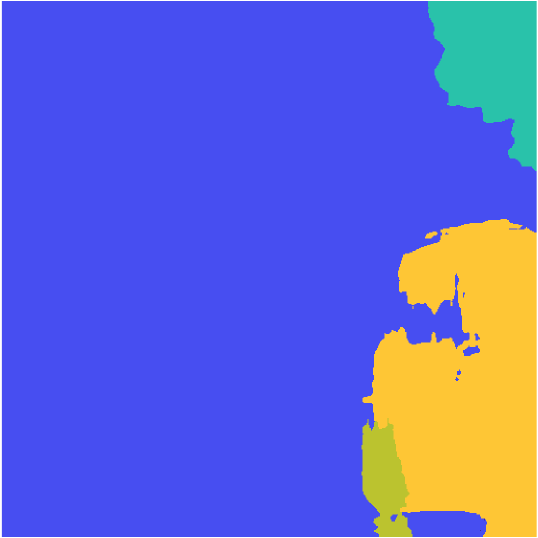}}   
    \end{minipage}%
    \begin{minipage}{0.25\linewidth}
	    \centering
	    {\includegraphics[width=\columnwidth]{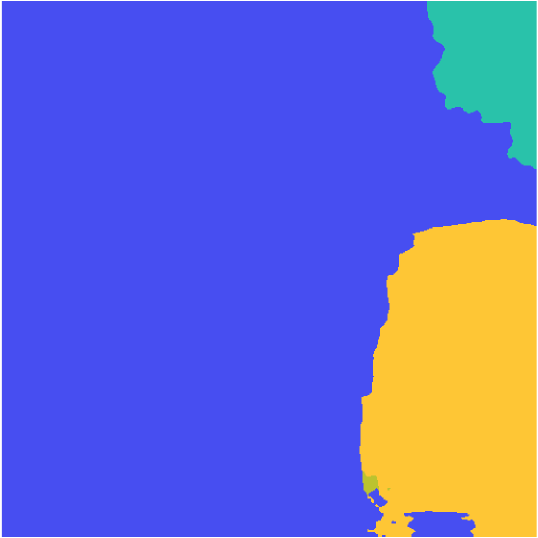}}   
    \end{minipage}%
    \begin{minipage}{0.25\linewidth}
	    \centering
	    {\includegraphics[width=\columnwidth]{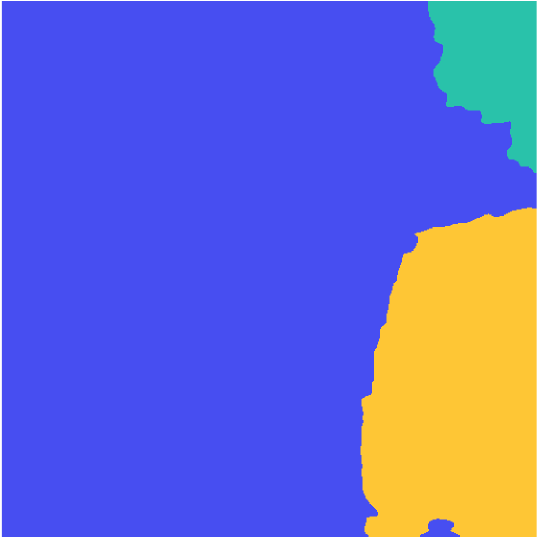}}   
    \end{minipage}%

    \begin{minipage}{0.25\linewidth}
	    \centering
	    {\includegraphics[width=\columnwidth]{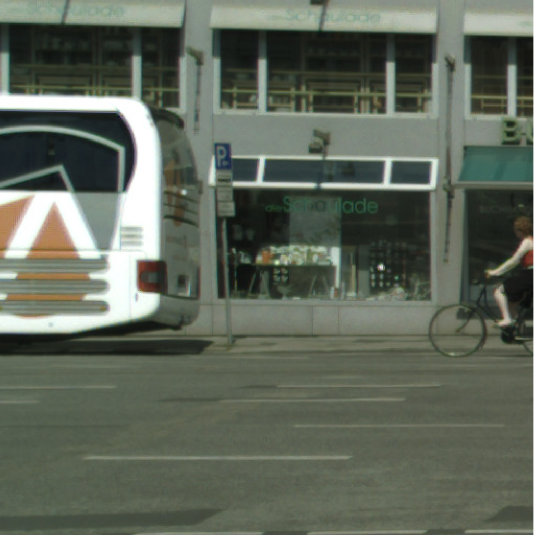}} 
    \end{minipage}%
    \begin{minipage}{0.25\linewidth}
	    \centering
	    {\includegraphics[width=\columnwidth]{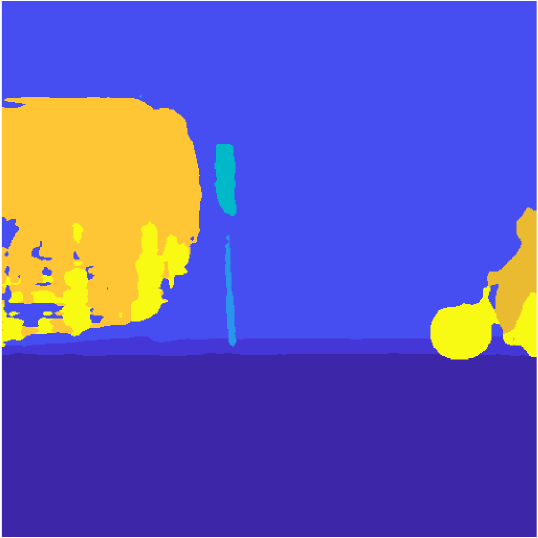}}   
    \end{minipage}%
    \begin{minipage}{0.25\linewidth}
	    \centering
	    {\includegraphics[width=\columnwidth]{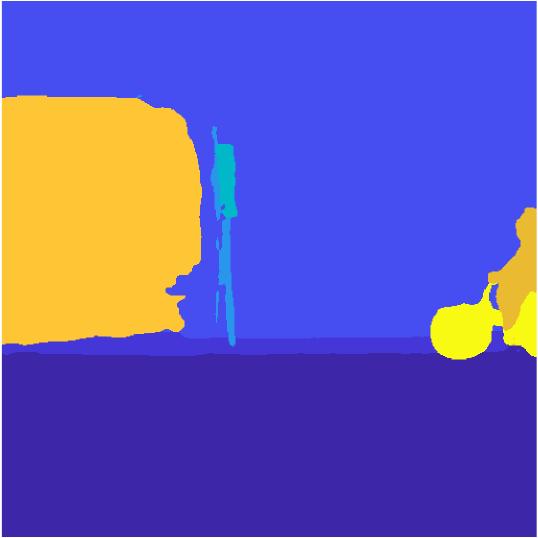}}   
    \end{minipage}%
    \begin{minipage}{0.25\linewidth}
	    \centering
	    {\includegraphics[width=\columnwidth]{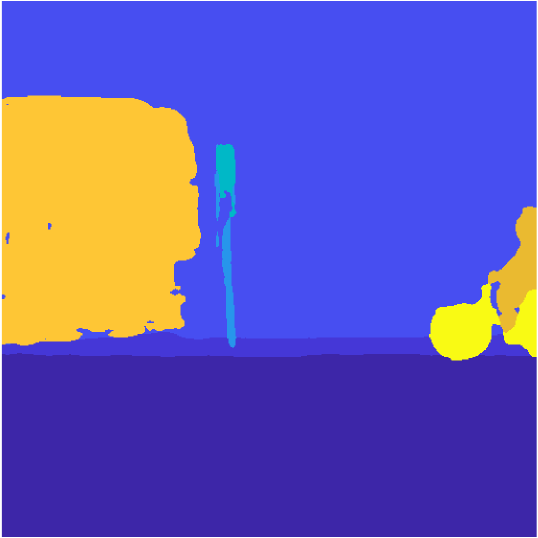}}   
    \end{minipage}%
    
    \begin{minipage}{0.25\linewidth}
	    \centering
	    \footnotesize Input
    \end{minipage}%
    \begin{minipage}{0.25\linewidth}
	    \centering
	    \footnotesize Baseline
    \end{minipage}%
    \begin{minipage}{0.25\linewidth}
	    \centering
	    \footnotesize Ours ($\times3$)
    \end{minipage}%
    \begin{minipage}{0.25\linewidth}
	    \centering
	    \footnotesize Ours ($\times9$)
    \end{minipage}%
    \vspace{-0.15cm}
	\caption{Qualitative results for semantic segmentation. }
    \vspace{-0.4cm}
	\label{fig:segment}
\end{figure}
\subsection{Task Generalization of the Proposed Method}
\noindent\textbf{1) Domain Generalization.} 
To evaluate robustness against domain gaps, we employed the PACS \cite{li2017deeper} dataset, which includes four distinct domains (art, cartoon, photo, and sketch). This dataset enables assessing model resilience to style disparities and few-shot scenarios. Like prior studies, we used a Leave-One-Domain-Out protocol, where we finetuned on three domains and tested on the remaining domain. We finetuned the ImageNet-pre-trained PVTv2 for 80 epochs. Table \ref{table:4} presents the results, demonstrating the robustness of our KNOW. In all domains, our KNOW consistently outperform the baseline, validating the effectiveness of our approach in adapting to diverse domain styles. These results underscore the generality and robustness of our enriched weights across different domains, highlighting the potential of our method for applications where adaptability and effective transfer learning are critical.

\noindent\textbf{2) Image Captioning.} 
We explored fine-tuning on a novel task with an unseen modality. We used the Flickr8K \cite{hodosh2013framing} dataset containing 8,000 images with five unique captions per image. These captions describe prominent entities and events within each image. For this, we reused the trained PVTv2 on ImageNet as the vision encoder backbone and attached a transformer-based text decoder initialized with random weights. We then finetuned the baseline weights and KNOW, on Flickr8K for 50 epochs using the Adam optimizer with data augmentation, measuring masked accuracy on the validation set. As shown in Table \ref{table:3}, ours improved the masked accuracy by approximately 2.2\%, indicating the robustness of our work even for novel modality. 

\noindent\textbf{3) Semantic Segmentation.}
Lastly, we applied our method to semantic segmentation using DeepLabV3+ \cite{chen2018encoder} with a MobileNet \cite{howard2017mobilenets} backbone. We incorporated KNOW predictions during pre-training on Pascal VOC \cite{Everingham15} with a ratio of $r=0.33$, without additional meta-training. For the downstream task, we finetuned on Cityscapes \cite{cordts2016cityscapes} for 150 epochs using Adam with cosine learning rate decay and augmentation, reporting mean class accuracy (mAcc) following the standard protocol. As shown in Table \ref{table:5}, task vector fails in the $\times 9$ case, performing worse than the baseline. For LogFit cases, it provides only marginal performance improvements. In contrast, our proposed method successfully enhances the mIoU values in both the $\times 3$ and $\times 9$ cases. These results demonstrate that our approach is effective even for more complex tasks beyond image classification. Additionally, several examples of segmentation results are provided in Fig. \ref{fig:segment}.
As shown, the baseline model often fails to detect traffic lights and produces unstable segmentation. In contrast, the proposed method ($\times 3$) addresses the instability problem, and the proposed method ($\times 9$) successfully generates well-defined segmentation results. These demonstrate that KNOW prediction improves downstream performance.

\section{Discussions}
\subsection{Ablation Study}
Our approach includes a hyperparameter \( S \), which we analyzed following the same protocol with ResNet18 and CIFAR100 for pre-training and CIFAR10 for downstream task outlined in Section \ref{sec:resnet}. We evaluated three values of \( S \in \{2,3,4,5\} \). Note that \( S = 1 \) can be considered as our baseline without weight prediction, while \( S = 2 \) is similar to the Task Vector approach. Thus, the cases of \( S \in \{2,3\} \) have already been evaluated in Table \ref{table:t1}. Using iterative prediction, we generated \([\hat{\Theta}^{-1}, \hat{\Theta}^{-2}, \hat{\Theta}^{-3}]\) (denoted as \(\times2\), \(\times4\), and \(\times8\), respectively) for each \( S \), and finetuned on CIFAR10. The results are summarized in Table \ref{table:6}.

\begin{table}[t]
    \renewcommand{\tabcolsep}{1.5mm}
    \centering
	\resizebox{1\linewidth}{!}{
		\begin{tabular}{l|ccc}
            \Xhline{3\arrayrulewidth}
            & \textbf{Pred.$(\times2)$} & \textbf{Pred.$(\times4)$} & \textbf{Pred.$(\times8)$} \\\hline
            $S=2$ (Task Vector) &$92.69\pm0.09$ &$92.70\pm0.09$& $92.65\pm0.13$ \\
            $S=3$ (KNOWN)& \bf 93.01 $\pm$ 0.16 & $93.04\pm0.06$ & $92.72\pm0.10$\\
            $S=4$ (KNOWN)& $92.97\pm0.16$ & \underline{93.10 $\pm$ 0.13} & \underline{92.89 $\pm$ 0.16}\\
            $S=5$ (KNOWN)& \underline{93.00 $\pm$ 0.11}& \bf 93.27 $\pm$ 0.09 & \bf 93.55 $\pm$ 0.05\\
            \Xhline{3\arrayrulewidth}
            \end{tabular}
        }
    \vspace{-0.25cm}
    \caption{Results of ablation study about $S$. Each value represents the test accuracy for CIFAR10}
    \vspace{-0.3cm}
    \label{table:6}
\end{table}
\subsection{Time Cost Analysis}
Our method requires a sequential forgetting process. Assuming a fixed batch size and number of iterations, this process incurs a time cost proportional to \((1 + r + r^2 + \dots + r^{S-2})\), where \( r \) is the sampling rate and falls within the range \( 0 < r < 1 \). Using the geometric sum formula for finite terms, this can be simplified to approximately \(\frac{1 - r^{S-1}}{1 - r}\) times the training duration. A smaller \( r \) reduces the required time cost while achieving greater knowledge enrichment. Notably, the time cost for weight prediction is negligible since it requires only inference without gradient computation. Also, our KNOWN is highly small with only 9,425 parameters, so it can be processed rapidly. Further, it is possible to batch-level processing. Therefore, predicting weights for ResNet18 takes less than $3.01\pm0.09$ seconds, indicating $2.6774\times 10^{-7}$ second for prediction of a parameter.

For example, we report the computational costs of experiments with ResNet18 and CIFAR100 in Section \ref{sec:resnet}. Table \ref{table:7} compares the time costs of the proposed method and the baseline. For the full training dataset (100\%), standard training of ResNet18 requires 38 seconds per epoch, with a total of 200 epochs for pre-training. As shown, the proposed method incurs a training cost of \(\frac{1 - r^{S-1}}{1 - r}\) times that of Na\"ive training due to progressive forgetting, while the prediction cost remains negligible. However, the accuracy gain surpasses that of the baseline trained with a dataset twice as large. Also, please note that the training cost can be reduced by adopting $r<0.5$. This result highlights the efficiency of the proposed method compared to practical data collection.

\begin{table}[t]
    \renewcommand{\tabcolsep}{0.5mm}
    \centering
	\resizebox{\linewidth}{!}{
		\begin{tabular}{c:c|c|cc|c}
            \Xhline{3\arrayrulewidth}
            \multicolumn{2}{c|}{\multirow{2}{*}[-0.em]{\bf Methods}} & \bf Dataset & \bf Training & \bf Inference & \bf Accuracy\\
             \multicolumn{2}{c|}{} &\bf Amount (\%) & \bf Time (s) & \bf Time (s) &  \bf (\%)\\
            \hline
            \multicolumn{2}{c|}{\multirow{2}{*}[-0.em]{\makecell{ Na\"ive Transfer\\ (Baseline)}}}  &50 & 3,800 & N/A & 92.08 \\
             \multicolumn{2}{c|}{} &100 & 7,600 & N/A & 92.40 \\\hline
            \multirow{3}{*}[-0.em]{\bf KNOW} & KNOWN ($\times2$)& 50 & 7,372 & 3.01 & 92.58 \\
             & KNOWN ($\times4$)& 50 & 7,372 & 6.02 & \underline{92.62} \\
             & KNOWN ($\times8$)& 50 & 7,372 & 9.03 & \textbf{93.11} \\\hline
            \Xhline{3\arrayrulewidth}
            \end{tabular}
        }
    \caption{Comparison in terms of time cost for KNOW prediction and test accuracy improvements for CIFAR10}
    \vspace{-0.3cm}
    \label{table:7}
    \vspace{-0.1cm}
\end{table}

\section{Conclusion}

In this paper, we address the challenge of limited training data by reinterpreting knowledge forgetting: \textit{intentionally inducing, then reversing it to recover and enrich the model’s knowledge.} 
Building on this insight, we propose \textbf{KNOW prediction}, a strategy that synthesizes knowledge-enhanced virtual weights, serving as an effective initialization that improves performance across diverse downstream tasks.
To further enhance, we developed \textbf{KNOWN}, a lightweight meta-trained hyper-model specifically designed to predict KNOW. Through extensive experiments, we demonstrate that the weights predicted by our method significantly improve performance across various downstream tasks, including image classification, captioning, domain generalization, and semantic segmentation. The feasibility of this approach is validated through in-depth analyses of weight trajectories, consistently showing improvements in training efficiency and transferability. Furthermore, when applied across different architectures and tasks without extra meta-training, KNOWN exhibits strong adaptability, achieving improvements even in novel domains and modalities. By providing knowledge-enriched weights, our approach offers a consistent advantage for diverse tasks and data-scarce scenarios, highlighting its potential for broader applications.

\section*{Acknowledgements}
This work was partly supported by the National Research Council  of Science \& Technology (NST) grant by the Korea government (MSIT)(No. GTL25041-000, 50\%), partly supported by the IITP-ITRC grant funded by the Korea government (MSIT)(IITP-2026-RS-2023-00258649, 25\%), and partly supported by IITP grant funded by the Korea government (MSIT)(IITP-2023-RS-2023-00266615, 25\%).

{
    \bibliographystyle{ieeenat_fullname}
    \bibliography{main}

@String(IJCV = {Int. J. Comput. Vis.})

@String(CVPR= {IEEE Conf. Comput. Vis. Pattern Recog.})

@String(ICCV= {Int. Conf. Comput. Vis.})

@String(ECCV= {Eur. Conf. Comput. Vis.})

@String(ICLR = {Int. Conf. Learn. Represent.})

@String(AAAI = {AAAI})

@String(IJCV  = {IJCV})

@String(CVPR  = {CVPR})

@String(ICCV  = {ICCV})

@String(ECCV  = {ECCV})

@String(ICLR  = {ICLR})

@article{kaplan2020scaling,
  title={Scaling laws for neural language models},
  author={Kaplan, Jared and McCandlish, Sam and Henighan, Tom and Brown, Tom B and Chess, Benjamin and Child, Rewon and Gray, Scott and Radford, Alec and Wu, Jeffrey and Amodei, Dario},
  journal={arXiv preprint arXiv:2001.08361},
  year={2020}
}

@inproceedings{sun2017revisiting,
  title={Revisiting unreasonable effectiveness of data in deep learning era},
  author={Sun, Chen and Shrivastava, Abhinav and Singh, Saurabh and Gupta, Abhinav},
  booktitle={ICCV},
  year={2017},
}

@inproceedings{gargiulo2025task,
  title={Task singular vectors: Reducing task interference in model merging},
  author={Gargiulo, Antonio Andrea and Crisostomi, Donato and Bucarelli, Maria Sofia and Scardapane, Simone and Silvestri, Fabrizio and Rodola, Emanuele},
  booktitle={CVPR},
  year={2025},
}

@inproceedings{wang2024localizing,
  title={Localizing task information for improved model merging and compression},
  author={Wang, Ke and Dimitriadis, Nikolaos and Ortiz-Jim{\'e}nez, Guillermo and Fleuret, Fran{\c{c}}ois and Frossard, Pascal},
  booktitle={ICML},
  year={2024},
}

@inproceedings{marczak2024magmax,
  title={Magmax: Leveraging model merging for seamless continual learning},
  author={Marczak, Daniel and Twardowski, Bart{\l}omiej and Trzci{\'n}ski, Tomasz and Cygert, Sebastian},
  booktitle={ECCV},
  year={2024},
}

@inproceedings{knyazev2024accelerating,
  title={Accelerating Training with Neuron Interaction and Nowcasting Networks}, 
  author={Boris Knyazev and Abhinav Moudgil and Guillaume Lajoie and Eugene Belilovsky and Simon Lacoste-Julien},
  booktitle={ICLR},
  year={2025}
}

@inproceedings{dauphin2019metainit,
  title={Metainit: Initializing learning by learning to initialize},
  author={Dauphin, Yann N and Schoenholz, Samuel},
  booktitle={NeurIPS},
  year={2019}
}

@inproceedings{knyazev2023can,
  title={Can we scale transformers to predict parameters of diverse imagenet models?},
  author={Knyazev, Boris and Hwang, Doha and Lacoste-Julien, Simon},
  booktitle={ICML},
  year={2023}
}

@inproceedings{he2019rethinking,
  title={Rethinking imagenet pre-training},
  author={He, Kaiming and Girshick, Ross and Doll{\'a}r, Piotr},
  booktitle={ICCV},
  year={2019}
}

@inproceedings{lopez2017gradient,
  title={Gradient episodic memory for continual learning},
  author={Lopez-Paz, David and Ranzato, Marc'Aurelio},
  booktitle={NeurIPS},
  year={2017}
}

@incollection{mccloskey1989catastrophic,
  title={Catastrophic interference in connectionist networks: The sequential learning problem},
  author={McCloskey, Michael and Cohen, Neal J},
  booktitle={Psychology of learning and motivation},
  year={1989}
}

@article{goodfellow2013empirical,
  title={An empirical investigation of catastrophic forgetting in gradient-based neural networks},
  author={Goodfellow, Ian J and Mirza, Mehdi and Xiao, Da and Courville, Aaron and Bengio, Yoshua},
  journal={arXiv preprint arXiv:1312.6211},
  year={2013}
}

@article{kirkpatrick2017overcoming,
  title={Overcoming catastrophic forgetting in neural networks},
    author = {James Kirkpatrick  and Razvan Pascanu  and Neil Rabinowitz  and Joel Veness  and Guillaume Desjardins  and Andrei A. Rusu  and Kieran Milan  and John Quan  and Tiago Ramalho  and Agnieszka Grabska-Barwinska  and Demis Hassabis  and Claudia Clopath  and Dharshan Kumaran  and Raia Hadsell },
  journal={Proceedings of the national academy of sciences},
  year={2017}
}

@article{huh2016makes,
  title={What makes ImageNet good for transfer learning?},
  author={Huh, Minyoung and Agrawal, Pulkit and Efros, Alexei A},
  journal={arXiv preprint arXiv:1608.08614},
  year={2016}
}

@article{raffel2020exploring,
  title={Exploring the limits of transfer learning with a unified text-to-text transformer},
  author={Raffel, Colin and Shazeer, Noam and Roberts, Adam and Lee, Katherine and Narang, Sharan and Matena, Michael and Zhou, Yanqi and Li, Wei and Liu, Peter J},
  journal={JMLR},
  year={2020}
}

@inproceedings{kornblith2019better,
  title={Do better imagenet models transfer better?},
  author={Kornblith, Simon and Shlens, Jonathon and Le, Quoc V},
  booktitle={CVPR},
  year={2019}
}

@inproceedings{abnar2021exploring,
  title={Exploring the limits of large scale pre-training},
  author={Abnar, Samira and Dehghani, Mostafa and Neyshabur, Behnam and Sedghi, Hanie},
  booktitle={ICLR},
  year={2022},
}

@inproceedings{shi2021overcoming,
  title={Overcoming catastrophic forgetting in incremental few-shot learning by finding flat minima},
  author={Shi, Guangyuan and Chen, Jiaxin and Zhang, Wenlong and Zhan, Li-Ming and Wu, Xiao-Ming},
  booktitle={NeurIPS},
  year={2021}
}

@article{luo2023empirical,
  title={An empirical study of catastrophic forgetting in large language models during continual fine-tuning},
  author={Luo, Yun and Yang, Zhen and Meng, Fandong and Li, Yafu and Zhou, Jie and Zhang, Yue},
  journal={IEEE Transactions on Audio, Speech and Language Processing},
  year={2025},
  volume={33},
  pages={3776-3786},
  publisher={IEEE}
}

@inproceedings{chen2019catastrophic,
  title={Catastrophic forgetting meets negative transfer: Batch spectral shrinkage for safe transfer learning},
  author={Chen, Xinyang and Wang, Sinan and Fu, Bo and Long, Mingsheng and Wang, Jianmin},
  booktitle={NeurIPS},
  year={2019}
}

@inproceedings{
jang2025learning,
title={Learning to Rewind via Iterative Prediction of Past Weights for Practical Unlearning},
author={Jang, Jinhyeok and Kim, Jaehong and Youn, Chan-Hyun},
booktitle={AAAI},
  year={2025}
}

@inproceedings{wortsman2022model,
  title={Model soups: averaging weights of multiple fine-tuned models improves accuracy without increasing inference time},
  author = {Wortsman, Mitchell and Ilharco, Gabriel and Gadre, Samir Ya and Roelofs, Rebecca and Gontijo-Lopes, Raphael and Morcos, Ari S and Namkoong, Hongseok and Farhadi, Ali and Carmon, Yair and Kornblith, Simon and Schmidt, Ludwig},
  booktitle={ICML 2022}
}

@inproceedings{ilharco2022patching,
  title={Patching open-vocabulary models by interpolating weights},
  author={Ilharco, Gabriel and Wortsman, Mitchell and Gadre, Samir Yitzhak and Song, Shuran and Hajishirzi, Hannaneh and Kornblith, Simon and Farhadi, Ali and Schmidt, Ludwig},
  booktitle={NeurIPS},
  year={2022}
}

@inproceedings{matena2022merging,
  title={Merging models with fisher-weighted averaging},
  author={Matena, Michael S and Raffel, Colin A},
  booktitle={NeurIPS},
  year={2022}
}

@inproceedings{draxler2018essentially,
  title={Essentially no barriers in neural network energy landscape},
  author={Draxler, Felix and Veschgini, Kambis and Salmhofer, Manfred and Hamprecht, Fred},
  booktitle={ICML},
  year={2018}
}

@inproceedings{garipov2018loss,
  title={Loss surfaces, mode connectivity, and fast ensembling of dnns},
  author={Garipov, Timur and Izmailov, Pavel and Podoprikhin, Dmitrii and Vetrov, Dmitry P and Wilson, Andrew G},
  booktitle={NeurIPS},
  year={2018}
}

@inproceedings{li2018visualizing,
  title={Visualizing the loss landscape of neural nets},
  author={Li, Hao and Xu, Zheng and Taylor, Gavin and Studer, Christoph and Goldstein, Tom},
  booktitle={NeurIPS},
  year={2018}
}

@inproceedings{golatkar2020eternal,
  title={Eternal sunshine of the spotless net: Selective forgetting in deep networks},
  author={Golatkar, Aditya and Achille, Alessandro and Soatto, Stefano},
  booktitle={CVPR},
  year={2020}
}

@inproceedings{ilharco2022editing,
  title={Editing models with task arithmetic},
  author={Ilharco, Gabriel and Ribeiro, Marco Tulio and Wortsman, Mitchell and Schmidt, Ludwig and Hajishirzi, Hannaneh and Farhadi, Ali},
  booktitle={ICLR},
  year={2022}
}

@inproceedings{jang2023learning,
  title={Learning to boost training by periodic nowcasting near future weights},
  author={Jang, Jinhyeok and Yun, Woo-han and Kim, Won Hwa and Yoon, Youngwoo and Kim, Jaehong and Lee, Jaeyeon and Han, ByungOk},
  booktitle={ICML},
  year={2023}
}

@InProceedings{Nilsback08,
   author = {Nilsback, M-E. and Zisserman, A.},
   title = {Automated Flower Classification over a Large Number of Classes},
   booktitle = {ICVGIP},
   year = {2008}
}

@article{wang2021pvtv2,
  title={Pvtv2: Improved baselines with pyramid vision transformer},
  author={Wang, Wenhai and Xie, Enze and Li, Xiang and Fan, Deng-Ping and Song, Kaitao and Liang, Ding and Lu, Tong and Luo, Ping and Shao, Ling},
  journal={Computational Visual Media},
  year={2022}
}

@inproceedings{
sinha2017introspectionaccelerating,
title={Introspection: Accelerating Neural Network Training By Learning Weight Evolution},
author={Abhishek Sinha and Aahitagni Mukherjee and Mausoom Sarkar and Balaji Krishnamurthy},
booktitle={ICLR},
  year={2017}
}

@inproceedings{deng2009imagenet,
  title={Imagenet: A large-scale hierarchical image database},
  author={Deng, Jia and Dong, Wei and Socher, Richard and Li, Li-Jia and Li, Kai and Fei-Fei, Li},
  booktitle={CVPR},
  year={2009}
}

@inproceedings{lv2017learning,
  title={Learning gradient descent: Better generalization and longer horizons},
  author={Lv, Kaifeng and Jiang, Shunhua and Li, Jian},
  booktitle={ICML},
  year={2017}
}

@inproceedings{wichrowska2017learned,
  title={Learned optimizers that scale and generalize},
  author={Wichrowska, Olga and Maheswaranathan, Niru and Hoffman, Matthew W and Colmenarejo, Sergio Gomez and Denil, Misha and Freitas, Nando and Sohl-Dickstein, Jascha},
  booktitle={ICML},   year={2017}
}

@inproceedings{andrychowicz2016learning,
  title={Learning to learn by gradient descent by gradient descent},
  author={Andrychowicz, Marcin and Denil, Misha and Colmenarejo, Sergio G{\'o}mez and Hoffman, Matthew W and Pfau, David and Schaul, Tom and Shillingford, Brendan and de Freitas, Nando},
  booktitle={NeurIPS},   year={2016}
}

@article{howard2017mobilenets,
  title={Mobilenets: Efficient convolutional neural networks for mobile vision applications},
  author={Howard, Andrew G and Zhu, Menglong and Chen, Bo and Kalenichenko, Dmitry and Wang, Weijun and Weyand, Tobias and Andreetto, Marco and Adam, Hartwig},
  journal={arXiv preprint arXiv:1704.04861},
  year={2017}
}

@inproceedings{cordts2016cityscapes,
  title={The cityscapes dataset for semantic urban scene understanding},
  author={Cordts, Marius and Omran, Mohamed and Ramos, Sebastian and Rehfeld, Timo and Enzweiler, Markus and Benenson, Rodrigo and Franke, Uwe and Roth, Stefan and Schiele, Bernt},
  booktitle={CVPR},   year={2016}
}

@Article{Everingham15,
   author = "Everingham, M. and Eslami, S. M. A. and Van~Gool, L. and Williams, C. K. I. and Winn, J. and Zisserman, A.",
   title = "The Pascal Visual Object Classes Challenge: A Retrospective",
   journal = "IJCV",
   year="2015"
}

@inproceedings{sandler2018mobilenetv2,
  title={Mobilenetv2: Inverted residuals and linear bottlenecks},
  author={Sandler, Mark and Howard, Andrew and Zhu, Menglong and Zhmoginov, Andrey and Chen, Liang-Chieh},
  booktitle={CVPR},   year={2018}
}

@inproceedings{resnet,
  title={Deep residual learning for image recognition},
  author={He, Kaiming and Zhang, Xiangyu and Ren, Shaoqing and Sun, Jian},
  booktitle={CVPR},   year={2016}
}

@inproceedings{kim2021robust,
  title={Robust small-scale pedestrian detection with cued recall via memory learning},
  author={Kim, Jung Uk and Park, Sungjune and Ro, Yong Man},
  booktitle={ICCV},
  year={2021}
}

@inproceedings{shufflenetv2,
  title={Shufflenet v2: Practical guidelines for efficient cnn architecture design},
  author={Ma, Ningning and Zhang, Xiangyu and Zheng, Hai-Tao and Sun, Jian},
  booktitle={ECCV},   year={2018}
}

@inproceedings{densenet,
  title={Densely connected convolutional networks},
  author={Huang, Gao and Liu, Zhuang and Van Der Maaten, Laurens and Weinberger, Kilian Q},
  booktitle={CVPR},   year={2017}
}

@article{wah2011caltech,
  title={The Caltech-UCSD Birds-200-2011 Dataset},
  author={Wah, Catherine and Branson, Steve and Welinder, Peter and Perona, Pietro and Belongie, Serge},
  year={2011},
  institution = {California Institute of Technology},
  number = {CNS-TR-2011-001}
}

@inproceedings{KrauseStarkDengFei,
title = {3D Object Representations for Fine-Grained Categorization},
booktitle = {ICCV Workshop},
year = {2013},
author = {Jonathan Krause and Michael Stark and Jia Deng and Li Fei-Fei}
}

@article{cifar10,
  title={Learning multiple layers of features from tiny images},
  author={Krizhevsky, Alex and Hinton, Geoffrey},
  year={2009},
  publisher={Citeseer}
}

@article{fashionmnist,
  title={Fashion-mnist: a novel image dataset for benchmarking machine learning algorithms},
  author={Xiao, Han and Rasul, Kashif and Vollgraf, Roland},
  journal={arXiv preprint arXiv:1708.07747},
  year={2017}
}

@misc{tiny_imagenet_challenge,
  title        = {Tiny ImageNet Challenge},
  author       = {{Stanford CS231N}},
  howpublished = {\url{https://cs231n.stanford.edu/tiny-imagenet-200.zip}},
  note         = {[Online; accessed 2026-03-16]}
}

@inproceedings{l2oamalgam,
  title={Optimizer Amalgamation},
  author={Huang, Tianshu and Chen, Tianlong and Liu, Sijia and Chang, Shiyu and Amini, Lisa and Wang, Zhangyang},
  booktitle={ICLR},
  year={2022}
}

@article{lecun1998gradient,
  title={Gradient-based learning applied to document recognition},
  author={LeCun, Yann and Bottou, L{\'e}on and Bengio, Yoshua and Haffner, Patrick},
  journal={Proceedings of the IEEE},
  year={1998}
}

@inproceedings{benton2021loss,
  title={Loss surface simplexes for mode connecting volumes and fast ensembling},
  author={Benton, Gregory and Maddox, Wesley and Lotfi, Sanae and Wilson, Andrew Gordon},
  booktitle={ICML},
  year={2021}
}

@inproceedings{yang2022towards,
  title={Towards theoretically inspired neural initialization optimization},
  author={Yang, Yibo and Wang, Hong and Yuan, Haobo and Lin, Zhouchen},
  booktitle={NeurIPS},
  year={2022}
}

@inproceedings{glorot2010understanding,
  title={Understanding the difficulty of training deep feedforward neural networks},
  author={Glorot, Xavier and Bengio, Yoshua},
  booktitle={AISTATS},   year={2010}
}

@inproceedings{he2015delving,
  title={Delving deep into rectifiers: Surpassing human-level performance on imagenet classification},
  author={He, Kaiming and Zhang, Xiangyu and Ren, Shaoqing and Sun, Jian},
  booktitle={ICCV},
  year={2015}
}

@inproceedings{zhu2021gradinit,
  title={Gradinit: Learning to initialize neural networks for stable and efficient training},
  author={Zhu, Chen and Ni, Renkun and Xu, Zheng and Kong, Kezhi and Huang, W Ronny and Goldstein, Tom},
  booktitle={NeurIPS},
  year={2021}
}

@inproceedings{li2017deeper,
  title={Deeper, broader and artier domain generalization},
  author={Li, Da and Yang, Yongxin and Song, Yi-Zhe and Hospedales, Timothy M},
  booktitle={ICCV},
  year={2017}
}

@article{hodosh2013framing,
  title={Framing image description as a ranking task: Data, models and evaluation metrics},
  author={Hodosh, Micah and Young, Peter and Hockenmaier, Julia},
  journal={JAIR},
  year={2013}
}

@inproceedings{jang2025model,
  title={Model stock: All we need is just a few fine-tuned models},
  author={Jang, Dong-Hwan and Yun, Sangdoo and Han, Dongyoon},
  booktitle={ECCV},
  year={2024}
}

@inproceedings{chen2018encoder,
  title={Encoder-decoder with atrous separable convolution for semantic image segmentation},
  author={Chen, Liang-Chieh and Zhu, Yukun and Papandreou, George and Schroff, Florian and Adam, Hartwig},
  booktitle={ECCV},
  year={2018}
}
}

\end{document}